\documentclass[conference]{IEEEtran}
\usepackage{paralist, tabularx}
\usepackage{graphicx}
\usepackage[dvipsnames,table,x11names]{xcolor}
\usepackage{hyperref}
\usepackage{booktabs} 
\usepackage{multirow}
\usepackage{amsmath}
\usepackage{amssymb}
\usepackage{amsfonts}

\newcommand{\mypar}[1]{\vspace{1pt}\noindent\textbf{#1.}}
\newcommand{\mypartwo}[1]{\vspace{1pt}\noindent\textit{#1.}}

\hypersetup{
	pdftoolbar=true,        
	pdfmenubar=true,        
	pdffitwindow=false,     
	pdfstartview={FitH},    
	pdftitle={Evaluating the Ability of LLMs to Solve Semantics-Aware Process Mining Tasks},    
	pdfauthor={Adrian Rebmann, Fabian David Schmidt, Goran Glavas, and Han van der Aa},     
	pdfsubject={},   
	pdfcreator={},   
	pdfproducer={}, 
	pdfkeywords={}, 
	pdfnewwindow=true,      
	colorlinks=true,       
	linkcolor=Brown,          
	citecolor=Brown,        
	filecolor=Brown,      
	urlcolor=Brown           
}

\begin{document}

\title{Evaluating the Ability of LLMs to Solve Semantics-Aware Process Mining Tasks}

\author{\IEEEauthorblockN{Adrian Rebmann\IEEEauthorrefmark{1}, Fabian David Schmidt\IEEEauthorrefmark{2}, Goran Glava{\v{s}}\IEEEauthorrefmark{2}, and Han van der Aa\IEEEauthorrefmark{3}} \IEEEauthorblockA{\IEEEauthorrefmark{1}Data and Web Science Group,
University of Mannheim, Germany\\
\textit{rebmann@uni-mannheim.de}} 
\IEEEauthorblockA{\IEEEauthorrefmark{2}Center for Artificial
Intelligence and Data Science,
University of Würzburg, Germany\\
\textit{\{fabian.schmidt, goran.glavas\}@uni-wuerzburg.de}}
\IEEEauthorblockA{\IEEEauthorrefmark{3}Faculty of Computer Science, University of Vienna, Austria\\
\textit{han.van.der.aa@univie.ac.at}}
}

\maketitle

\begin{abstract}
The process mining community has recently recognized the potential of large
language models (LLMs) for tackling various process mining tasks. 
Initial studies report the capability 
of LLMs to support process analysis and even, to some extent, that they are able to reason about how processes work.
This latter property suggests that LLMs could
also be used to tackle process mining tasks that benefit from an understanding of process behavior. Examples of such tasks include (semantic) anomaly detection and next activity prediction, which both involve considerations of the meaning of activities  and their inter-relations. 
In this paper, we investigate the capabilities of LLMs to tackle such semantics-aware process mining tasks.
Furthermore, whereas most works on the intersection of LLMs and process mining only focus on testing these models out of the box,
we provide a more principled investigation of the  utility of LLMs for process mining, including their ability to obtain process mining knowledge post-hoc by means of in-context learning and supervised fine-tuning.
Concretely, we define three process mining tasks that benefit from an understanding of process semantics and provide extensive benchmarking datasets for each of them.
Our evaluation experiments reveal that 
 (1) LLMs fail to solve challenging process mining tasks out of the box and when provided only a handful of in-context examples, (2) but they yield strong performance  when fine-tuned for these tasks, consistently surpassing smaller, encoder-based language models.

\end{abstract}


\IEEEpeerreviewmaketitle

\section{Introduction}
\label{sec:introduction}

Process mining involves analyzing event data from organizational processes to gain actionable insights into the true manner in which they are executed. Recently, the process mining community has taken notice of the potential of large language models (LLMs) in tackling process mining tasks. Exploratory studies have looked into the effectiveness of these generative models for this purpose, yielding promising initial results. 
Among others, LLMs demonstrate the ability to describe processes textually using event data, answer questions about the event data, and assist in formulating queries to address specific analysis questions~\cite{berti2023abstractions, jessen2023chit, torres2024mapping}. 
This capability shows their potential to offer valuable assistance to process analysts in manual data exploration and analysis. 

The greatest potential from an automated process analysis point-of-view lies in LLMs' impressive natural language understanding capabilities. 
These capabilities could provide a strong foundation for supporting process mining tasks, particularly those that benefit from a consideration of the meaning of process steps and their relations. 
Examples include anomaly detection~\cite{van2021natural} and next activity prediction~\cite{neu2022systematic}. 
Anomaly detection can improve performance by identifying undesired process behavior based on the meaning of activities. For instance, it can detect when a delivery is created despite the corresponding purchase order being canceled.
Next activity prediction can enhance performance by narrowing down the set of potential next activities to those that make semantic sense. For example, it can discard a \emph{check request} activity if the request has already been approved.

The capabilities of (smaller) language models for tackling process mining tasks such as semantic anomaly detection have been shown~\cite{caspary2023does}. However, an investigation of the utility of LLMs for solving such semantic-aware process mining tasks is still missing. This particularly relates to in-depth evaluations of the capabilities of such models, e.g., with respect to their ability to obtain process mining knowledge post-hoc by means of in-context learning and supervised fine-tuning.
A reason for this is the lack of properly defined natural language processing (NLP) tasks that effectively conceptualize the capability 
to perform semantics-aware process mining tasks.
Consequently, there is also a lack of benchmarking datasets similar to those available for established NLP tasks such as Question Answering~\cite{rajpurkar2016squad} and Text Summarization~\cite{narayan2018don}, hindering structured evaluation experiments.


In this paper, we address these problems through the following contributions:

\begin{compactitem}
\item We define three tasks (\autoref{sec:tasks}) that allow for assessing the capabilities of (large) language models for semantics-aware process mining: 
(1) determining whether an activity sequence represents a valid execution of a process,
(2) deciding whether the execution order of two activities is valid, and 
(3) selecting the next process activity to be performed given an incomplete activity sequence.

\item  We provide a corpus of process behavior (\autoref{sec:datasets}) based on the largest publicly available process model collection that can be used to establish benchmarking datasets for semantic process analysis tasks. Based on this corpus, we establish and provide benchmarking data sets that allow for conducting quantitative evaluations of the performance of language models on the three proposed tasks.

\item We assess the performance of open-source LLMs to solve the proposed tasks based on the established benchmarking datasets in an experimental evaluation (\autoref{sec:setup}--\autoref{sec:results}). Therein, we compare LLMs in \emph{in-context learning} and \emph{fine-tuning} settings and discriminative encoder-based language models.

\end{compactitem}
Our results indicate that the proposed tasks are too challenging for LLMs to solve them out of the box and using in-context learning. 
However, if an LLM is fine-tuned on a specific task it can achieve accurate results and consistently outperforms discriminative encoder-based language models. 

\section{Preliminaries}
\label{sec:preliminaries}
In this section, we define preliminaries on process models, event data, and eventually-follows relations,  which we require for the remainder of the paper.

\mypar{Process Models}
Let $\mathcal{A}$ be the universe of possible activities that can be performed in organizational processes. 
We define a process model $M$ as the set of executions that are allowed in a process. 
Each execution $\pi$ is represented as an activity sequence $\pi = \langle a_1,...,a_n \rangle$, with $a_i \in \mathcal{A}$. We use $A_M \subseteq \mathcal{A}$, to denote the set of activities that appear in the sequences of $M$.

\mypar{Event Data} We adopt a simple event model, focusing on the control-flow of a process. A trace $\sigma$ is a sequence that represents the events that have been recorded for the  execution of a single instance of an organizational process. Such a trace consists of a finite sequence of activities $\sigma = \langle a_1,...,a_n\rangle$,  with $a \in \mathcal{A}$. An event log $L$ is a finite multi-set of traces. $A_L \subseteq \mathcal{A}$ denotes the set of activities that appear in the traces of $L$.

\mypar{Eventually-Follows Relations} 
We use the eventually-follows relation $\prec$ to capture ordering relations between pairs of activities (stemming either from the traces of an event log or from activity sequences of a process model). 
Given a trace $\sigma = \langle a_1,..., a_n \rangle$, we use $a_i \prec_\sigma a_j$ to denote that $a_i$ occurs (directly or indirectly) before $a_j$ in the trace, with $1\leq i < j \leq n$. 
Accordingly, $a_i \prec_M a_j$ holds if a model $M$ allows for an activity sequence $\pi$ in which activity $a_i$ occurs before $a_j$. $\mathit{EF}_M$ denotes all eventually-follows relations of the activity sequences allowed by $M$.

\section{Tasks}
\label{sec:tasks}
This section describes and defines three process mining tasks that benefit from an understanding of process behavior. We design these tasks (1) such that there is no access to historical event data and (2) to consider solely the control-flow perspective.
(1) allows us to assess whether a language model can solve the tasks based purely on its encoded knowledge of how processes generally work. (2) ensures that a language model determines its results exclusively based on the meaning of the activities and their interrelations, excluding other aspects like data attributes, which are often highly organization-specific. 
Our tasks include two forms of  \emph{semantic anomaly detection}, as well as \emph{semantic next activity prediction}.

\subsection{Semantic Anomaly Detection}
Anomaly detection in process mining aims to identify outlying process behavior in the traces of an event log~\cite{van_der_aalst_process_2005}.
Many approaches do this by identifying statistical outliers~\cite{bezerra_algorithms_2013}, arguing that behavior is anomalous if it is infrequent. 

By contrast, \textit{semantic} anomaly detection~\cite{van2021natural} focuses on the identification of process behavior that does not make sense. Arguing that, just because behavior is infrequent, does not make it anomalous, whereas just because something happens regularly, does not mean that it is proper process behavior.
For instance, from a semantic point of view, an invoice should not be created if the corresponding purchase order has been rejected and should therefore be detected as anomalous behavior, independent of how frequently this happens.

Detecting anomalies based on process semantics requires a different approach compared to frequency-based anomaly detection.  
Whereas frequency-based detection can be performed by just using data in an event log (revealing statistical outliers), semantic anomaly detection requires information about how a process should (or should not) work in general. By definition, such information needs to be obtained from outside of the event log, e.g., from large knowledge bases or---as we do in this paper---from the knowledge encoded in LLMs.

We define two specific tasks in this context, focusing on the trace and activity-relation levels:

\mypar{Trace-Level Semantic Anomaly Detection} 
Trace-level semantic anomaly detection (T-SAD) is a binary classification problem in which a trace $\sigma \in L$ needs to classified as anomalous or not, according to its semantics. 
For instance, given a trace $\sigma=\langle$\emph{register application}, \emph{approve application}, \emph{review application}$\rangle$, the task is to classify that $\sigma$ is anomalous. This is the case because an application should first be reviewed and only then approved (or rejected).
The challenge here is that there is no specification available of the process at hand that can be used for this. Rather that anomality needs to be inferred, requiring an understanding of how processes generally work.

\mypar{Activity-Level Semantic Anomaly Detection}
Activity-level semantic anomaly detection (A-SAD) is a more fine-granular task than T-SAD, focusing on pairs of activities in a trace rather than on an entire trace at once.
Specifically, A-SAD focuses on classifying any eventually-follows relation $ a_i \prec_\sigma a_j$ of two activities $a_i$ and $a_j$ that appear in a respective order in a trace $\sigma$ as being anomalous or not. 
For instance, given the trace $\sigma=\langle$\emph{create purchase order}, \emph{reject purchase order}, \emph{create invoice}$\rangle$, the eventually-follows relation \emph{reject purchase order} $\prec_\sigma$ \emph{create invoice} should be classified as anomalous, whereas the other pairwise relations, i.e.,
\emph{create purchase order} $\prec_\sigma$ \emph{reject purchase order} and \emph{create purchase order} $\prec_\sigma$ \emph{create invoice} should be classified as valid.

\subsection{Semantic Next Activity Prediction}
Next activity prediction, also known as \emph{next event} or \emph{next step} prediction, is a key predictive process monitoring task. 
Its objective is to determine the subsequent activity in an ongoing process execution~\cite{neu2022systematic}. Various approaches, predominantly based on supervised deep learning (e.g.,~\cite{evermann2017predicting, pfeiffer2021multivariate}), have been developed to tackle this task. 

As a semantics-aware counterpart for next activity prediction, we introduce the semantic next activity prediction (S-NAP) task. 
For an incomplete trace $\sigma$, which represents an ongoing process execution in which $k$ activities have been performed ($k\geq1$), the task is to predict the next activity $a_{k+1}$ in $\sigma$ based on a set of possible activities $A$. 
For instance, given $\sigma = \langle$\emph{create purchase order}, \emph{approve purchase order}$\rangle$ and $A=\{$\emph{create purchase order}, \emph{approve purchase order}, \emph{create invoice}, \emph{make payment}$\}$, the task is to predict $a_{k+1}$ as \emph{create invoice}. This is because, generally, an invoice should be created before a payment is made. 

Whereas approaches for (traditional) next activity prediction train a model on historical traces from an event log $L$ to predict the next activity in ongoing (i.e., unseen) executions of the process, S-NAP focuses on situations where no such historical traces are available. As a result, the next activity must be inferred by considering the semantics of the activities involved in a process.

\section{Datasets}
\label{sec:datasets}
This section details the creation and characteristics of the text corpus and benchmarking datasets that we use to evaluate the ability of language models to solve the proposed tasks.
We make all datasets publicly available~\cite{rebmann_2024_11276246}.

\subsection{A Corpus of Process Behaviors}
Language models require textual input. In order to assess their ability to solve semantics-aware process mining tasks we, therefore, need a collection of textual representations of process behavior, a so-called \emph{corpus}.
This corpus then serves as a basis to create task-specific data that can be used for training and evaluating language models on the proposed tasks. 

Since no such corpus is readily available, we create one based on graphical process models (i.e., process diagrams).
To this end, we use \textsc{sap-sam}~\cite{sola2023sap}, which is the largest publicly available collection of process diagrams to date. 
In order to create a high-quality corpus, we select only English BPMN diagrams from \textsc{sap-sam} that meet specific requirements. These ensure that the corpus includes only unique and valid process behavior.
In particular, we require that a diagram can be transformed into a sound workflow net and that no two diagrams have the same activity set. 
The former requirement mitigates data quality issues in the \textsc{sap-sam} collection~\cite{sola2023sap} and ensures that we can properly generate activity sequences from the diagram. The latter makes sure that we do not include duplicate behavior. 
Furthermore, we require a diagram to contain at least two different activities to ensure that it actually captures ordering relations between different activities.

We use the workflow net of each selected diagram to generate activity sequences, capturing all executions allowed by the net. 
For loops, we ensure that each loop is executed at most once, so that we capture relations involving rework, yet, obtain a finite set of activity sequences. 
For each net, this yields a process model $M$ according to the definition in \autoref{sec:preliminaries}, i.e., a set of activity sequences capturing its allowed behavior. 
We add each such $M$ to the corpus.

We show the characteristics of the resulting corpus in \autoref{tab:corpus}. As depicted there, the complexity of the process models varies considerably. For instance, the number of unique activities is 4.9 on average, whereas the maximum is 21 and the process models allow for 10.34 activity sequences on average, whereas the maximum amount is 10,080.

\begin{table}[!htb]
    \centering
    \caption{Characteristics of the process behavior corpus.}
    \label{tab:corpus}
    \begin{tabular}{lrrrr}
    \toprule
     \textbf{Characteristic} & \textbf{Total} & \multicolumn{3}{c}{\textbf{Per process model}} \\
     & & \textbf{Avg.} & \textbf{Min.} &  \textbf{Max.}\\
    \midrule
    \textbf{\# Process models} &15,857 & --& --&--\\
    \textbf{\# Unique activities} & 49,108 & 4.70 &2 &21\\
    \textbf{\# Unique activity sequences} &163,484 & 10.34& 1& 10,080\\
    \bottomrule
    \end{tabular}
\end{table}

Based on this corpus, we can generate task-specific datasets for the individual tasks.

\subsection{Task-Specific Benchmarking Datasets}
With a text corpus of process behaviors available, we can generate task-specific benchmarking datasets that include task samples and a gold standard. This gold standard enables objective, quantitative evaluation of the language models on the tasks based on established evaluation measures.
The characteristics of the dataset are shown in \autoref{tab:benchmark}, whereas they are established in the manner described in the following.

\begin{table}[!htb]
    \centering
    \caption{Characteristics of the task-specific benchmarking datasets.}
    \label{tab:benchmark}
    \begin{tabular}{lrrr}
    \toprule
     \textbf{Task Dataset} & \textbf{Total} & \textbf{Valid} & \textbf{Anomalous}\\
    \midrule
    \textbf{T-SAD} & 291,251 &150,301 & 140,950\\
    \textbf{A-SAD} & 316,308& 158,154& 158,154 \\
    \textbf{S-NAP}& 1,289,081&  1,289,081 &- \\
    \bottomrule
    \end{tabular}
\end{table}

\mypar{T-SAD} 
To establish the T-SAD dataset, we first create an event log $L$ for each process model $M$ in the corpus such that each $\pi \in M$ becomes a trace $\sigma \in L$. To make sure that there is a minimum number of traces per log, we randomly duplicate traces in $L$ until a size of 100 is reached if $L$ does not already contain at least 100 traces. 
Subsequently, for each trace $\sigma \in L$, we make a decision regarding the insertion of noise, with a 50 percent probability. 
This noise insertion involves swapping two randomly selected activities within the trace. 
After swapping, we check whether the resulting sequence $\sigma'$ is indeed anomalous, i.e., $\sigma' \notin M$. 
If the sequence is found to still be valid, i.e., $\sigma' \in M$, we continue iterating through potential swaps until we obtain an anomalous sequence\footnote{We limit the number of retries to 10 per trace to guarantee termination.}. 
This ensures that the dataset contains (roughly) the same amount of valid and anomalous traces, which is crucial for robust model training and evaluation. 

Each of the 291,251 records of the T-SAD dataset then consists of a trace $\sigma$, the correct label of $\sigma$, i.e.,  $\mathit{Anomalous}$ if $\sigma \notin M$ and $\mathit{Valid}$ otherwise, and the set of possible activities in the process from which $\sigma$ originates as context information.

\mypar{A-SAD} We create the A-SAD dataset based on the set of eventually-follows relations $\mathit{EF}_M$ for each process model $M$ in the corpus. 
The relations in $\mathit{EF_M}$ represent all valid execution orders of activities of $M$.
Next to these, we create a set of anomalous relations $\mathit{EF_{\not M}}$ , i.e., ones that are not in $\mathit{EF_M}$. 
To provide a balanced dataset, we establish $\mathit{EF_{\not M}}$ by randomly selecting relations that are not in $\mathit{EF_M}$, until we have an equal number of valid and anomalous relations.

Each of the 316,308 records of the  A-SAD dataset consists of an eventually-follows relation $r$, the correct label of $r$, i.e., $\mathit{Anomalous}$ if $r \notin EF_M$ and $\mathit{Valid}$ otherwise, and the set of all activities $A_M$ in the process model from which the activities in $r$ originate as context information.

\mypar{S-NAP} For the S-NAP dataset, we first create an event log $L$ for each process model $M$ in the corpus such that each $\pi \in M$ becomes a trace $\sigma \in L$.
Then, we generate all possible prefixes for each trace $\sigma \in L$ and add them to $L$. 
This involves iteratively considering sub-traces of increasing length $k$ from the first activity of a trace $\sigma$, up to its full length, which ensures that every potential prefix is captured. 

Each of the 1,289,081 records of the S-NAP dataset then consists of a length-$k$ prefix ($\sigma_k$) of $\sigma$, the correct label of $\sigma_k$, i.e., the activity at position $k+1$ in $\sigma$, and the set of possible activities $A_L$ of the event log from which $\sigma$ originates as context information.

\section{LLM-Based Process Mining}
\label{sec;llms}
Neural language models, based on the Transformer architecture \cite{vaswani_attention_2017,devlin2019bert} come in two main flavors: (1) bidirectional language models, also commonly called \textit{encoders}, which are typically (pre-)trained via masked language modeling objectives in which masked tokens are predicted from both left and right context \cite{devlin2019bert,liu2019roberta} and (2) unidirectional language models, also known as \textit{decoders}, which are trained via autoregressive language modeling objectives where the next token is predicted from the preceding context \cite{brown2020language,touvron2023llama}. 
LLMs are very large instances of the latter category (with at least a billion parameters) that are, following large-scale autoregressive language modeling, typically additionally trained for \textit{instruction following}, i.e., to provide solutions to tasks given the natural language description of these tasks \cite{wang2022super,zhang2023instruction}. 
Such instruction-tuning allows LLMs to generalize to new tasks through textual task descriptions (since capturing meaning of language is what LLMs excel at) and solve them successfully even when not provided with any task-specific (training) examples.

\mypar{Fine-Tuning LMs on Classification Tasks}
Fine-tuning is the process of further training a pretrained language model, in order to specialize it for a specific (classification or regression) task. 
The advantage compared to training a model from scratch is that the training data size for fine-tuning is considerably smaller, thus reducing resources required to train a task-specific model. We fine-tune language models for each of our semantics-aware process mining tasks, all of which are classification tasks. We next briefly describe the fine-tuning procedures for (a) discriminative classification with an encoder LM and (b) generative classification with a decoder LM.    

\begin{figure}[!t]
    \centering
    \includegraphics[scale=0.63]{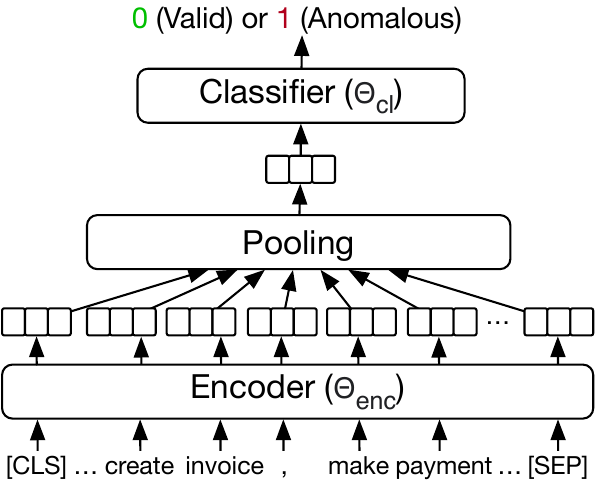}
    \caption{Illustration of discriminative classification with an \textit{encoder} LM.}
    \label{fig:encoder}
    \vspace{-0.5em}
\end{figure}

\mypartwo{Discriminative Fine-Tuning of Encoder LMs}
To fine-tune an encoder LM for classification tasks, we extend the model's base architecture (i.e., the pretrained Transformer network) with an additional classification layer: the parameters of the classifier are trained from scratch (i.e., randomly initialized), whereas the encoder's parameters are updated (i.e., fine-tuned).
Fine-tuning of an encoder LM for the T-SAD task is illustrated in \autoref{fig:encoder} using a record with a trace $\langle$\emph{create invoice}, \emph{make payment}, $\dots \rangle$ as input. The input is first split into a sequence of subword tokens.\footnote{For more frequent words in the language, a token will commonly correspond to the whole word; less frequent words, on the other hand, will often be broken down into more frequent subtokens (e.g., \textit{``tokenization''} may be segmented into \textit{``token''} and \textit{``ization''}). The exact subword vocabulary is model dependent, i.e., each LM comes with its own tokenizer.} 
The actual input for encoder LMs is commonly surrounded with synthetic sequence start (\texttt{[CLS]}) and sequence end (\texttt{[SEP]}) tokens. 
The encoder (i.e., the Transformer network) outputs one vector---a transformed/contextualized representation---for each token in the input sequence, including the sequence start/end tokens. Let $\mathbf{x}_{CLS} \in \mathbb{R}^d$ be the representation of the sequence start token \texttt{CLS} (output of the encoder) with $d$ as the hidden size of the encoder's Transformer network; this vector $\mathbf{x}_{CLS}$ can be seen as a latent semantic representation of the whole input text and is forwarded as input to the classifier. The classifier, in turn, is a single-layer feed-forward network: $\hat{\textbf{y}} = \mathit{softmax}(\mathbf{W}_{cl} \cdot \mathbf{x}_{CLS} + \mathbf{b}_{cl})$; $\mathbf{W}_{cl} \in \mathbb{R}^{c \times d}$ and $\mathbf{b}_{cl} \in \mathbb{R}^{c}$ are the trainable parameters of the classifier ($c$ is the number of classes in the classification task) and $\mathit{softmax}$ is the function commonly used to convert real-valued vectors into probability distributions---the final output $\hat{\textbf{y}}$ is thus a probability distribution over the task's classes. We train the model (jointly update the parameters of both classifier and encoder in end-to-end fashion) by minimizing the widely used cross-entropy loss, i.e., the negative logarithm of the probability that the model predicted for the true class of the input instance.

T-SAD and A-SAD are binary classification tasks (i.e., $c = 2$) in which the model predicts whether the traces and ordered activity pairs, respectively, are \emph{Valid} or \emph{Anomalous}. S-NAP is a multi-class classification task in which the set of classes is defined with the activities in $A_M$ of the process model $M$ from which the input record was created.

\mypartwo{Generative Fine-Tuning of Decoder (L)LMs}
Autoregressively trained decoder LLMs cast classification tasks as language generation tasks. Concretely, each class into which the preceding text is to be classified is assigned one token from the vocabulary and the LLM's language modeling head (a classifier over the LLM's vocabulary) is supposed to generate the token of the correct class. 
For example, for the T-SAD task, we convert individual training instances into prompts that couple (1) the set of process activities with (2) the concrete trace (or activity sequence) that is to be judged as \textit{Valid} or \textit{Anomalous}. Then we append the prompt that asks whether the sequence is anomalous, with the token \textit{true} assigned to the \textit{Anomalous} sequences and token \textit{false} to the \textit{Valid} sequences. The whole input for the decoder LLM for a single sequence is shown below (the label token is underlined and in blue):   

\smallskip
\noindent \emph{Activities: \{create order, approve order, reject order, create invoice, make payment\}} \\
\emph{Activity sequence: [create order, reject  order, create invoice, make payment]}\\
\emph{Anomalous: \color{blue}{\underline{true}}}

\begin{figure}[!t]
    \vspace{-0.5em}
    \centering
    \includegraphics[scale=0.63]{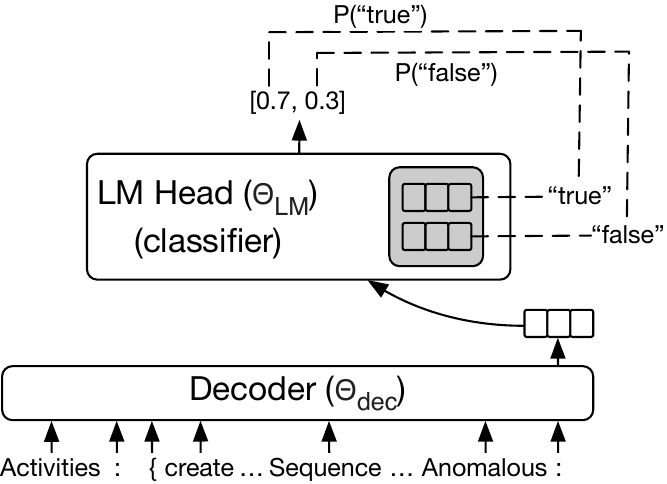}
    \caption{Illustration of constrained generative fine-tuning of a \textit{decoder} LM.}
    \label{fig:decoder}
    \vspace{-0.5em}
\end{figure}

\smallskip
We fine-tune a decoder LLM via \textit{constrained text generation}: given the entire preceding context (everything except the last token that indicates the class), we predict the next token, but allow the language modeling head to only predict the probabilities for the \textit{allowed} class tokens (in the above example, only {\textit{true}, \textit{false}}), as opposed to LLM pretraining in which the next token is predicted over the entire vocabulary of the LLM. We illustrate constrained generative fine-tuning of a decoder LM in \autoref{fig:decoder}.     
The Transformer network of the decoder produces the output representation by contextualizing all preceding tokens; the resulting vector is next compared against the representations of the allowed class tokens (in the example, {\textit{true}, \textit{false}}) to produce scores that are then converted into probabilities using softmax. We minimize the negative log likelihood of the probability assigned to the correct class token: as updating all LLM parameters is computationally infeasible, we resort to parameter-efficient fine-tuning via low-rank adaptation (LoRA) \cite{hu2021lora}.\footnote{For brevity, we refer the reader to the original work for details on LoRA.}

\mypar{Few-Shot In-Context Learning}
In-context learning (ICL) aims to induce a model to perform a task by providing a small set of input-label examples (so-called ``shots'') along with the task description; the query sample---the example (input) for which the label is to be generated---is provided at the end of the \textit{prompt}~\cite{dong2022survey}. In-context learning allows the LLM to understand the task (via the description and a few labeled examples), without supervised fine-tuning (i.e., without any updates to LLM's parameters).

\begin{figure}[!t]
\vspace{-0.5em}
    \centering
    \includegraphics[scale=0.5]{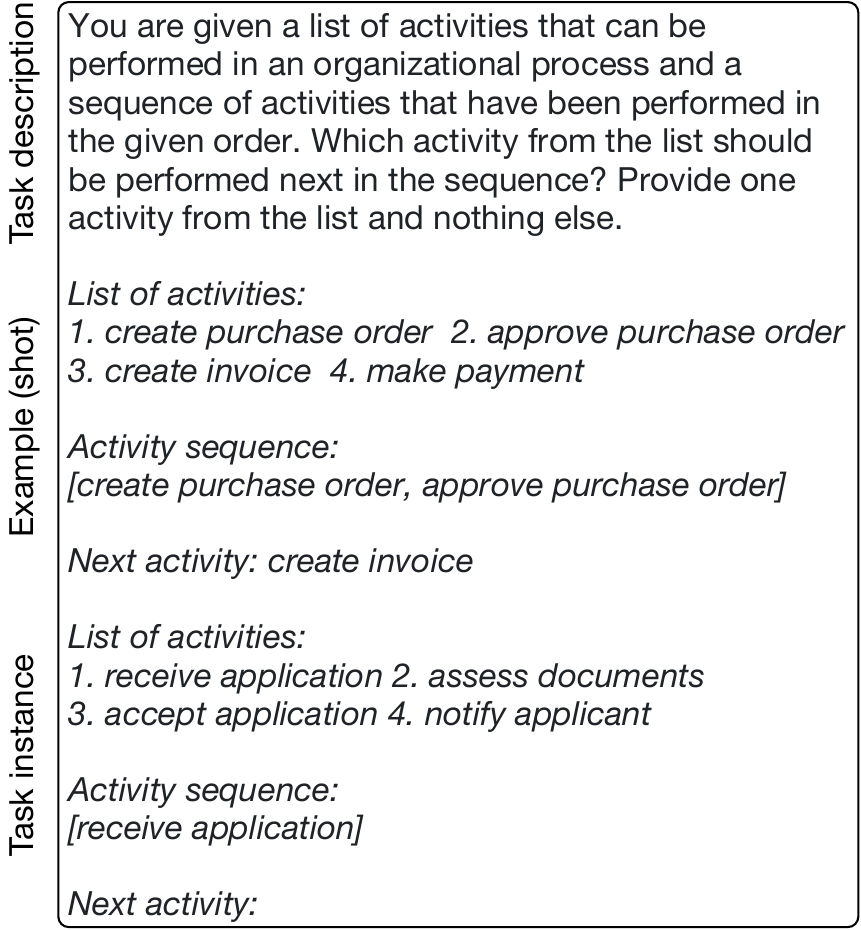}
    \caption{One-shot in-context-learning prompt for the S-NAP task.}
    \label{fig:snap_prompt}
\vspace{-0.5em}
\end{figure}

\autoref{fig:snap_prompt} displays the one-shot prompt for the S-NAP task: the description of the task is followed by one labeled task instance---list of possible activities and the prefix trace in question, along with the correct label---and then the query instance for which the LLM is to generate the label. 

\section{Experimental Setup}
\label{sec:setup}
We first describe how we split the datasets for training, validation, and testing, and then introduce the concrete LMs we use. We then provide details on our ICL and fine-tuning setups. For reproducibility, we make all training and evaluation scripts publicly available.\footnote{\url{https://github.com/a-rebmann/llms4pm}}

\mypar{Dataset Portions}
We split all datasets based on the process models from which the samples originate using 70\% of instances for training, 20\% for validation, and 10\% for final performance evaluation. 
For these splits, we ensure that no activity sequence of a model in the training sets appears in any of the process models in validation and test portions. This ensures that no knowledge of the process behavior is leaked from training to the validation and test sets, allowing us to correctly assess the generalization abilities of the LMs. Furthermore, we ensure a comparable complexity distribution of process models across splits via stratified sampling based on the number of unique activities in the models. \autoref{tab:split} displays the sizes of all splits for all three of our tasks. 

\begin{table}[!htb]
\vspace{-0.5em}
    \centering
    \caption{Training, validation, and test split characteristics per task.}
    \label{tab:split}
    \begin{tabular}{lrrrr}
    \toprule
     \textbf{Task} & \textbf{Total} & \textbf{Train} & \textbf{Validation}& \textbf{Test} \\
    \midrule
    \textbf{T-SAD} &291,251 &227,892 &43,609&19,750 \\
    \textbf{A-SAD} & 316,308& 229,402& 56,154 & 30,752\\
    \textbf{S-NAP}& 1,289,081&1,071,529&166,811 & 50,741\\
    \bottomrule
    \end{tabular}
\vspace{-0.5em}
\end{table}

\mypar{Large Language Models}
We resort to two widely used open decoder-LLMs that have demonstrated impressive performance on NLP benchmarks, namely: (1) \emph{Llama-3} (\emph{Llama}) LLM in its 8 billion parameter version\footnote{\url{https://huggingface.co/meta-llama/Meta-Llama-3-8B}} and (2) \emph{Mistral-2} (\emph{Mistral}) in its 7 billion parameter version\footnote{\url{https://huggingface.co/mistralai/Mistral-7B-Instruct-v0.2}}. We evaluate both in few-shot ICL and fine-tuning setups.  

\mypar{Baselines}
We use two baselines in our experiments: (1) We compare LLMs with ICL (i.e., without task-specific fine-tuning) against a random classifier, which assigns task's classes to test instances with equal probability; (2) We compare generatively fine-tuned LLMs against a discriminatively fine-tuned encoder LM: specifically, we use \emph{RobERTa}~\cite{liu2019roberta} in its \emph{large} version,\footnote{\url{https://huggingface.co/FacebookAI/roberta-large}} a strong and widely used English-only bidirectional encoder LM.

\mypar{Performance Metric} 
We measure performance using the macro F$_1$-score, so that classes equally contribute to the performance regardless of their size. Macro F$_1$ is the simple average of per-class F$_1$ scores, with F$_1$ of class $c$ being the harmonic mean between $c$'s precision and recall.       

\mypar{In-Context Learning and Prompt Optimization}
We experimented with several task formulation prompts for each task and selected the optimal one based on validation performance. We evaluated 6-shot, 10-shot, and 20-shot ICL for all three tasks: for binary T-SAD and A-SAD tasks, we evenly balance positive and negative instances; for S-NAP we sample one instance from randomly chosen 6, 10, or 20 process models, respectively. 
Different task description prompts led to marginal performance differences. Somewhat surprisingly, prompts with fewer shots produced better validation performance: we thus finally evaluate 6-shot ICL on our test data.\footnote{See \url{https://github.com/a-rebmann/llms4pm} for the task prompts we used.}

\mypar{Fine-Tuning}
We fine-tune \emph{Llama} and \emph{Mistral} in batches of two instances with gradient accumulation over 16 batches, resulting in an effective batch size of 32. We fine-tune \emph{RoBERTa} also in batches of 32 instances. All models are trained using the AdamW algorithm \cite{loshchilov2018decoupled}, with an initial learning rate of 1e-5. We fine-tune the LLMs for three epochs and \emph{RoBERTa} for ten epochs.  
We run each combination of task and model three times using different random seeds, corresponding to different random initialization of model parameters and shuffling of training data in each run.

\section{Results and Discussion}
\label{sec:results}
We report the performance of LLMs on the three tasks, followed by an in-depth analysis of the models' predictions and a discussion on training effort.  

\subsection{Main Results}
\label{sec:results:single}
\autoref{tab:results:single} shows the main results of our experiments per task for the random baseline, the two LLMs (\emph{Llama} and \emph{Mistral}) using the ICL approach, as well as \emph{RoBERTa} and the LLMs using the fine-tuning approach (FT). 
We report mean and ($\pm$) standard deviation variance over three runs for ICL and three different random seeds for FT. 
We first discuss the much weaker ICL results and then focus on fine-tuning performance. 

\begin{table}[!htb]
\vspace{-0.5em}
    \centering
    \caption{Main Results (macro $F_1$ scores).}
    \label{tab:results:single}
\begin{tabular}{lccc}
\toprule
\multirow{2}{*}{\textbf{Approach}}& & \textbf{Task} & \\
 & \textbf{T-SAD}& \textbf{A-SAD} & \textbf{S-NAP}\\
\midrule
\textbf{Random} & 0.50 $\pm$ 0.000 & 0.50 $\pm$ 0.000 & 0.13 $\pm$ 0.000 \\
\textbf{ICL Mistral} & 0.49 $\pm$ 0.022 & 0.44 $\pm$ 0.011 & 0.18 $\pm$ 0.018 \\
\textbf{ICL Llama} & 0.51 $\pm$ 0.015 & 0.53 $\pm$ 0.021 & 0.32 $\pm$ 0.054 \\ \midrule
\textbf{FT RoBERTa} & 0.77 $\pm$ 0.006 & 0.85 $\pm$ 0.003 & 0.63 $\pm$ 0.048 \\ 
\textbf{FT Mistral} & \textbf{0.79} $\pm$ 0.010 & \textbf{0.88} $\pm$ 0.002 & 0.68 $\pm$ 0.039 \\
\textbf{FT Llama} & \textbf{0.79} $\pm$ 0.011 & \textbf{0.88} $\pm$ 0.000 & \textbf{0.69} $\pm$ 0.049 \\
\bottomrule
\end{tabular}
\vspace{-0.5em}
\end{table}

\mypar{In-Context Learning Results}
For ICL, we find that the performance of the LLMs is at best marginally better (\emph{Llama}) at worst  (\emph{Mistral}) slightly worse than random performance for the two semantic anomaly detection tasks, T-SAD and A-SAD. Specifically, \emph{Llama} achieves a macro $F_1$-score of 0.51 for T-SAD and 0.53 for A-SAD, while \emph{Mistral} scores 0.44 resp.\ 0.49. These results indicate that the LLMs have not effectively learned these tasks from the few examples provided in the context. For the S-NAP task, ICL with LLMs does outperform the random baseline, with \emph{Llama} exhibiting much stronger performance (19-point gain over the random baseline) than \emph{Mistral} (only 5-point gain). The performance is nonetheless fairly poor in absolute terms (mere 0.32 with \textit{Llama}).  
These results suggest that these process mining tasks drastically differ from the language processing tasks on which the LLM instruction-tuning was carried out. 

\mypar{Fine-Tuning Results}
Poor ICL performance shows that LLMs \textit{a priori} know very little about process semantics and thus need to be explicitly trained for our tasks. The fine-tuned encoder-LM baseline, i.e., \emph{RoBERTa}, already  achieves drastically better performance than ICL with LLMs: for example, it obtains the F$_1$-score of 0.77 on the T-SAD task, which is an improvement of massive 26 points over the best ICL performance (0.51 by \emph{Llama}).
Fine-tuning the LLMs yields even better performance, with \emph{Llama} and \emph{Mistral} achieving an $F_1$-score of 0.79, a further 2-point improvement over \emph{RoBERTa}'s performance. 
The same trend holds for the other two tasks: 
on A-SAD, both \emph{Mistral} and \emph{Llama} yield very strong performance $F_1$-scores of 0.88, outperforming \emph{RoBERTa} by 3 points; on S-NAP, \emph{Llama} and \emph{Mistral} achieve $F_1$-scores of 0.69 and 0.68, respectively (6- and 5-point respective gains over \emph{RoBERTa}).

These results show that decoder-based LLMs can effectively acquire the missing process knowledge through explicit task-specific fine-tuning, yielding better results than their (smaller) encoder-based counterparts such as \emph{RoBERTa}. 
The fine-tuned LLMs consistently outperform \emph{RoBERTa} on all three tasks, which points to the benefits of much larger-scale pretraining to which they have been comparatively exposed.

\mypar{Task Comparison}
The results also demonstrate considerable differences in difficulty between the tasks. 
For A-SAD, the LLMs achieve an impressive $F_1$-score of 0.88, while the maximum score for T-SAD is 0.79, and the best model scores only 0.69 for S-NAP. This aligns with expectations. Solving T-SAD requires the model to identify whether process behavior is valid within the context of an entire trace, whereas A-SAD only necessitates assessing a single behavioral relation.
The S-NAP task is by far the most challenging of the three and is simply unsolvable for many instances. For example, consider a process that allows for the concurrent execution of activities. In such cases, it is indeterminable---for both humans and automated approaches---which activity occurs next in a trace based solely on a prefix, as there are multiple valid options. 

\subsection{In-Depth Analysis}
\label{sec:results:indepth}
To determine if the models can handle certain types of processes better than others, we conducted an in-depth analysis of the classification results obtained from the fine-tuned language models. For brevity, we focus on T-SAD, as its medium complexity best represents the three tasks.

We find that both the encoder and LLMs accurately detect anomalous traces across a wide range of process domains. However, they appear to be particularly effective at identifying anomalies for standard process types. For example, in a claim-handling process, they correctly identify anomalies such as $\langle$\emph{Enter and verify claim}, \emph{Handle payment}, \emph{Assess claim}$\rangle$, where the claim should be assessed before sending a payment. They also correctly detect that the trace $\langle$\emph{Confirm order}, \emph{Ship product}, \emph{Get shipment address}, \emph{Emit Invoice}, \emph{Receive Payment}$\rangle$ is anomalous given that the product is shipped before the address is determined in this order-handling process. 

For more specialized processes, we observe that LLMs often outperform RoBERTa. For instance, \emph{Llama} correctly identifies the trace $\langle$\emph{Arrival}, \emph{Treatment}, \emph{Triage}, \emph{Discharge}, \emph{Invoicing}$\rangle$ of a hospital process as anomalous, since \emph{Triage} should occur before \emph{Treatment}. In contrast,  \emph{RoBERTa} fails to detect this anomaly. Conversely, \emph{RoBERTa} incorrectly flags the trace $\langle$\emph{Disassemble system}, \emph{Refurbish materials}, \emph{Clean and paint covers}, \emph{Mount materials}, \emph{Move to bay}, \emph{Calibrate}, \emph{Handover}$\rangle$ of a refurbishing process as anomalous, even though it is valid, whereas \emph{Llama} correctly classifies this trace as valid. 

Finally, there are instances where both the LLMs and \emph{RoBERTa} incorrectly identify valid traces as anomalous. For example, both \emph{Llama} and \emph{RoBERTa} flag the trace $\langle$\emph{Receive invoices of partners}, \emph{Handle payment of customer}, \emph{Receive review}, \emph{Send payment to partners}$\rangle$ as anomalous, even though it is valid. According to the corresponding process model in the corpus, a review can be received at any point during an execution of the process, making this trace valid. However, this specificity might also be challenging for a human to determine without further contextual information.

\subsection{Training Effort}
\label{sec:results:effort}
Finally, we consider the effort required to train the language models on the tasks. 
Since in-context learning does not require any training effort, as the task-specific knowledge is provided at inference time, we focus on the training effort of fine-tuning an LLM (\emph{Llama}) versus an encoder (\emph{RoBERTa}).

\autoref{tab:results:effort} shows the run times for fine-tuning the language models for the different tasks per epoch, i.e., pass over all training samples.\footnote{Note that the models were trained on the same type of GPU.} 
As shown, the LLM requires considerably more time for training than the encoder baseline. In particular, training \emph{Llama} on the tasks takes up to 25 times longer than training \emph{RoBERTa} per epoch on the same data.
For instance, while fine-tuning \emph{Llama} for A-SAD takes 15 hours, \emph{RoBERTa} requires only around 40 minutes per epoch. 
This difference can be attributed to the huge number of parameters that need to be updated for the LLM during fine-tuning, even when using parameter-efficient fine-tuning. However, it is worth stressing that the LLM requires considerably fewer epochs to converge in terms of validation loss across tasks. This indicates that it not only learns the tasks better (as shown in the previous subsections), but also with fewer passes over the training data. 

\begin{table}[!htb]
\vspace{-0.5em}
    \centering
    \caption{Average run times for fine-tuning (per epoch).}
    \label{tab:results:effort}
    \begin{tabular}{lrrr}
\toprule
\multirow{2}{*}{\textbf{Approach}}& \multicolumn{3}{c}{\textbf{Task}} \\
 & \textbf{T-SAD} & \textbf{A-SAD} & \textbf{S-NAP}  \\
\midrule
\textbf{FT RoBERTa} & 0.5h & 0.6h & 1.3h  \\
\textbf{FT Llama} & 11.1h & 15.0h & 23.0h  \\
\bottomrule
\end{tabular}
    \vspace{-0.5em}
\end{table}

\section{Related Work}
\label{sec:related}
Neural language models have been used for various process analysis tasks, e.g., for annotating event logs with semantic information~\cite{rebmann_enabling_2022},  detecting anomalies~\cite{caspary2023does}, and constructing event logs based on textual records of process steps~\cite{kecht2021event}.

With the success of LLMs, several studies have explored their efficacy in handling process analysis tasks~\cite{torres2024mapping}. These tasks include transforming textual process descriptions into formal process models \cite{grohs2023large, kourani2024process}, generating textual descriptions from process data such as models and event logs~\cite{berti2023abstractions}, and identifying potential bottlenecks and undesired process behavior~\cite{berti2023abstractions}. However, much of the existing research relies on closed-source GPT models or proprietary software like \emph{ChatGPT}, limiting structured and reproducible evaluations~\cite{torres2024mapping}.

The lack of proper evaluations of LLMs for process mining tasks has recently been discussed in the process mining community~\cite{torres2024mapping, berti2024evaluating}.
As a response, Berti et al.\ proposed a benchmark\footnote{\url{https://github.com/fit-alessandro-berti/pm-llm-benchmark}} for process mining analysis questions. It consists of 52 prompts that are used to query various LLMs, whose answers are then rated by GPT-4o. Although the benchmark provides interesting insights, using an LLM to rate the results can yield biased outcomes. Such bias arises, e.g., from the tendency of LLMs to recognize and favor their own output~\cite{panickssery2024llm}.

In contrast, we define process mining tasks that benefit from an understanding of process behavior and evaluate LLMs using extensive task-specific benchmarking datasets in both, in-context learning and fine-tuning settings. Furthermore, our datasets provide gold standards, which allows for using established evaluation measures for classification tasks, eliminating the need for a proxy LLM to assess output quality.

\section{Conclusion}
\label{sec:conclusion}
In this paper, we investigated the capabilities of LLMs to solve semantics-aware process mining tasks, i.e., tasks that benefit from an understanding of process semantics. 
We defined three such tasks and provide an extensive benchmarking dataset for each of them.
Our evaluation experiments that use them show that LLMs fail to solve these challenging process mining tasks out of the box and in a few-shot in-context setting. 
However, our results demonstrate that LLMs achieve accurate performance when fine-tuned for these tasks, surpassing smaller, encoder-based language models.

In the future, we want to investigate the integration of state-of-the-art process mining approaches with LLMs. Since we have shown that LLMs can solve semantics-aware process mining tasks through encoded knowledge of process semantics, integrating existing process mining approaches with LLMs may yield performance improvements for classical process mining tasks they address. 
For example, an existing next-activity prediction approach could be extended by an LLM-based semantic check that rejects predictions that do not make sense, thereby improving the overall prediction performance.

\section*{Acknowledgment}
We received support by the state of Baden-Württemberg through bwHPC.

\medskip
\mypartwo{Reproducibility}
\emph{Our training and evaluation scripts are available through the project repository linked in \autoref{sec:setup}. Our process behavior corpus and benchmarking datasets are published separately}~\cite{rebmann_2024_11276246}.

\bibliographystyle{IEEEtran}
\bibliography{IEEEabrv, refs}

\end{document}